\def\year{2022}\relax
\documentclass[letterpaper]{article} 
\usepackage{aaai22}  
\usepackage{times}  
\usepackage{helvet}  
\usepackage{courier}  
\usepackage[hyphens]{url}  
\usepackage{graphicx} 
\usepackage{natbib}  
\usepackage{caption} 
\DeclareCaptionStyle{ruled}{labelfont=normalfont,labelsep=colon,strut=off} 
\usepackage{algorithm}
\usepackage{algorithmic}
\usepackage{smile}
\usepackage{wrapfig}

\usepackage{newfloat}
\usepackage{listings}
\floatstyle{ruled}
\newfloat{listing}{tb}{lst}{}
\floatname{listing}{Listing}
\title{Efficient Robust Training via Backward Smoothing}
\author {
    Jinghui Chen\textsuperscript{\rm 1},
    Yu Cheng\textsuperscript{\rm 2},
    Zhe Gan\textsuperscript{\rm 2},
    Quanquan Gu\textsuperscript{\rm 3},
    Jingjing Liu\textsuperscript{\rm 4}
}
\affiliations {
    \textsuperscript{\rm 1} Pennsylvania State University\\
    \textsuperscript{\rm 2} Microsoft Corporation\\
    \textsuperscript{\rm 3} University of California, Los Angeles\\ \textsuperscript{\rm 4} Tsinghua University \\
    {jzc5917@psu.edu,
    \{yu.cheng, zhe.gan\}@microsoft.com,
    qgu@cs.ucla.edu,  JJLiu@air.tsinghua.edu.cn}
}

\usepackage{bibentry}

\begin{document}

\maketitle

\begin{abstract}
Adversarial training is so far the most effective strategy in defending against adversarial examples. However, it suffers from high computational costs due to the iterative adversarial attacks in each training step. Recent studies show that it is possible to achieve fast Adversarial Training by performing a single-step attack with random initialization. However, such an approach still lags behind state-of-the-art adversarial training algorithms on both stability and model robustness. In this work, we develop a new understanding towards Fast Adversarial Training, by viewing random initialization as performing randomized smoothing for better optimization of the inner maximization problem. Following this new perspective, we also propose a new initialization strategy, \emph{backward smoothing}, to further improve the stability and model robustness over single-step robust training methods.
Experiments on multiple benchmarks demonstrate that our method achieves similar model robustness as the original TRADES method while using much less training time ($\sim$3x improvement with the same training schedule). 
\end{abstract}



\section{Introduction}\label{sec:intro}
Deep neural networks are well known to be vulnerable to adversarial examples \citep{szegedy2013intriguing}, \emph{i.e.}, a small perturbation on the original input can lead to misclassification or erroneous prediction. Many defense methods have been developed to mitigate the disturbance of adversarial examples \citep{guo2017countering,xie2017mitigating,song2017pixeldefend,ma2018characterizing,samangouei2018defense,dhillon2018stochastic,madry2017towards,zhang2019theoretically}, among which robust training methods, such as adversarial training \citep{madry2017towards} and TRADES \citep{zhang2019theoretically}, are currently the most effective strategies.
Specifically, adversarial training method \citep{madry2017towards} trains a model on adversarial examples by solving a min-max optimization problem:
\begin{align}
    \min_{\btheta} \frac{1}{n}\sum_{i=1}^n \max_{\xb_i' \in \cB_{\epsilon}(\xb_i)} L(f_{\btheta}(\xb_i'), y_i),
\end{align}
where $\{(\xb_i, y_i)\}_{i=1}^n$ is the training dataset, $f(\cdot)$ denotes the logits output of the neural network, $\cB_{\epsilon}(\xb_i):= \{\xb: \|\xb - \xb_i\|_\infty \leq \epsilon\}$ denotes the $\epsilon$-perturbation ball, and $L$ is the cross-entropy loss. 

On the other hand, instead of directly training on adversarial examples, TRADES \citep{zhang2019theoretically} further improves model robustness with a trade-off between natural accuracy and robust accuracy, by solving the empirical risk minimization problem with a robust regularization term:
\begin{align}
    \min_{\btheta} &\frac{1}{n}\sum_{i=1}^n \Big[ L(f_{\btheta}(\xb_i), y_i) \notag  \\
    &+\beta\max_{\xb_i' \in \cB_{\epsilon}(\xb_i)} \text{KL}\big(s(f_{\btheta}(\xb_i)), s(f_{\btheta}(\xb_i'))\big)\Big],
\end{align}
where $s(\cdot)$ denotes the softmax function, and $\beta>0$ is a regularization parameter. The goal of this robust regularization term (\emph{i.e.}, KL divergence term) is to ensure the outputs are stable within the local neighborhood. 
Both adversarial training and TRADES achieve good model robustness, as shown on recent model robustness leaderboards\footnote{\url{https://github.com/fra31/auto-attack} and \url{https://github.com/uclaml/RayS}.} \citep{croce2020reliable,chen2020rays}. However, a major drawback lies in that both are highly time-consuming for training, limiting their usefulness in practice. This is largely due to the fact that both methods perform iterative adversarial attacks (\emph{i.e.}, Projected Gradient Descent) to solve the inner maximization problem in each outer minimization step.

Recently, \cite{Wong2020Fast} shows that it is possible to use single-step adversarial attacks to solve the inner maximization problem, which previously was believed impossible. The key ingredient in their Fast AT approach is adding a random initialization step before the single-step adversarial attack. This simple change leads to a reasonably robust model that outperforms other fast robust training techniques, \emph{e.g.}, \cite{shafahi2019adversarial}. However, the simple change also has its downsides: 1) random initialization makes single-step robust training possible yet it can be quite unstable \citep{li2020towards}; 2) compared to state-of-the-art robust training models \citep{madry2017towards,zhang2019theoretically}, Fast AT still lags behind on model robustness. 
Besides these, It also remains a mystery in \cite{Wong2020Fast} on why random initialization is empirically effective. 

Although some attempts have been made trying to explain the role of random initialization and further improve Fast AT \citep{andriushchenko2020understanding,li2020towards}, in this work, we aim to understand the role of random initialization in \cite{Wong2020Fast} from a new perspective and further improve the model robustness-efficiency trade-off over previous fast robust training methods. Specifically, We propose a new principle towards understanding Fast AT - that random initialization can be viewed as performing randomized smoothing for better optimization of the inner maximization problem. 
In order to further improve the robustness-efficiency trade-off of fast robust training techniques, we propose a new initialization strategy, \emph{backward smoothing}, which strengthens the smoothing effect within the $\epsilon$-perturbation ball. 
The resulting method significantly improves both stability and model robustness over the single-step random initialization strategies. Moreover, even comparing with full-step robust training methods such as TRADES \citep{zhang2019theoretically}, our proposed backward smoothing strategy achieves similar model robustness while consuming much less training time ($\sim3$x improvement with the same training schedule). 

The remainder of this paper is organized as follows: in Section \ref{sec:related}, we briefly review existing literature on adversarial attacks, robust training as well as randomized smoothing technique. We present our new understanding of random initialization in Section \ref{sec:why}. We present our proposed method in Section \ref{sec:method}. In Section \ref{sec:exp}, we empirically evaluate our proposed method with other state-of-the-art baselines. Finally, we conclude this paper in Section \ref{sec:con}. 
 
\section{Related Work}\label{sec:related}
There exists a large body of literature on adversarial attacks and defenses. In this section, we only review the most relevant work to ours.

\noindent\textbf{Adversarial Attack}\,
The concept of adversarial examples was first proposed in \cite{szegedy2013intriguing}. Since then, many methods have been proposed, such as Fast Gradient Sign Method (FGSM) \citep{goodfellow6572explaining}, and Projected Gradient Descent (PGD)  \citep{kurakin2016adversarial, madry2017towards}. Later on, various attacks \citep{papernot2016limitations,moosavi2016deepfool,carlini2017towards,athalye2018obfuscated,chen2018frank,croce2020minimally,sriramanan2020guided,tashiro2020diversity} were also proposed for better effectiveness or efficiency.
 
There are also many attacks focused on different attack settings. \cite{chen2017zoo} proposed a black-box attack where the gradient is not available, by estimating the gradient via finite-differences. Various methods \citep{ilyas2018black,AlDujaili2020Sign,moon2019parsimonious,andriushchenko2020square, tashiro2020diversity} have been developed to improve the query efficiency of \cite{chen2017zoo}. Other methods \citep{brendel2018decisionbased,cheng2018queryefficient,cheng2020signopt} focused on the more challenging hard-label attack setting, where only the prediction labels are available.
On the other hand, there is recent work \citep{croce2020reliable,chen2020rays} that aims to accurately evaluate the model robustness via an ensemble of attacks or effective hard-label attack.

\noindent\textbf{Robust Training}\, 
Many heuristic defenses  \citep{guo2017countering,xie2017mitigating,song2017pixeldefend,ma2018characterizing,samangouei2018defense,dhillon2018stochastic} were proposed when the concept of adversarial examples was first introduced. However, they are later shown by \cite{athalye2018obfuscated} as not truly robust.
Adversarial training \citep{madry2017towards} is the first effective method towards defending against adversarial examples. Various adversarial training variants \citep{wang2019convergence,Wang2020Improving, zhang2019theoretically,wu2020adversarial,sriramanan2020guided,zhang2020attacks} were later proposed to further improve the adversarially trained model robustness. A line of researches focus on studying various others factors affecting model robustness such as early-stopping \citep{rice2020overfitting}, model width \citep{wu2021wider}, loss landscape \citep{liu2020loss} and parameter tuning \citep{pang2021bag,gowal2020uncovering}.
Another line of research utilizes extra information (\emph{e.g.}, pre-trained models \citep{hendrycks2019using} or extra unlabeled data \citep{carmon2019unlabeled,alayrac2019labels}) to further improve robustness. 

Recently, many focus on improving the training efficiency of adversarial training based  algorithms, such as free adversarial training \citep{shafahi2019adversarial} and Fast AT \citep{Wong2020Fast}, which uses single-step attack (FGSM) with random initialization. \cite{li2020towards} proposed a hybrid approach for improving Fast AT which is orthogonal to ours. \cite{andriushchenko2020understanding} proposed a new regularizer promoting gradient alignment for more stable training. 
Yet, its model robustness still falls behind the state-of-the-arts.

\noindent\textbf{Randomized Smoothing}\, \cite{duchi2012randomized} proposed the randomized smoothing technique and proved variance-based convergence rates for non-smooth
optimization. Later on, this technique was applied to certified adversarial defenses \citep{cohen2019certified,salman2019provably} for building robust models with certified robustness guarantees. In this paper, we are not targeting certified defenses. Instead, we use the randomized smoothing concept in optimization to explain Fast AT.

\section{Pros and Cons of Random Initialization}\label{sec:why}
In this section, we analyze the pros and cons of random initialization in Fast AT \citep{Wong2020Fast}. First, let us explain why random initialization in Fast AT is effective by looking into why one-step AT would fail without random initialization. 

\vspace{-2pt}
\subsection{What Caused the Failure of One-step AT Without Random Initialization?}

\cite{Wong2020Fast} has already shown that without random initialization, one-step AT would almost surely fail in the training procedure due to catastrophic overfitting, i.e., the robust accuracy w.r.t. a PGD adversarial suddenly drops to near $0$ even on training data. However, it is not clear what exactly causes this phenomenon. One natural conjecture is that perhaps the one-step attack is not effective enough for adversarial training purposes. Recall that the perturbation is obtained by solving the following inner maximization problem in adversarial training:
\begin{align}\label{eq:inner_max}
     \bm{\delta}^* = \argmax_{\bm{\delta}
     \in \cB_{\epsilon}(\zero)} L(f_{\btheta}(\xb + \bm{\delta}), y).
\end{align}
To figure out whether the attack effectiveness is the key cause for the poor performance of the plain one-step AT without random initialization, we conduct the following simple experiments by observing the loss increment after attack in each training step, i.e., $$\Delta_L = L(f_{\btheta}(\xb + \bm{\delta}^*), y) - L(f_{\btheta}(\xb), y),$$
where $\{(\xb, y)\}$ is the clean training example and $\bm{\delta}^*$ is the solution from \eqref{eq:inner_max}.
Since \eqref{eq:inner_max} aims at maximizing the loss value, this loss increment term $\Delta_L$ should always be positive along the entire training trajectory. 

\begin{figure}[t!]
  \centering
  \setlength{\belowcaptionskip}{-15pt}
  \includegraphics[width=.4\textwidth]{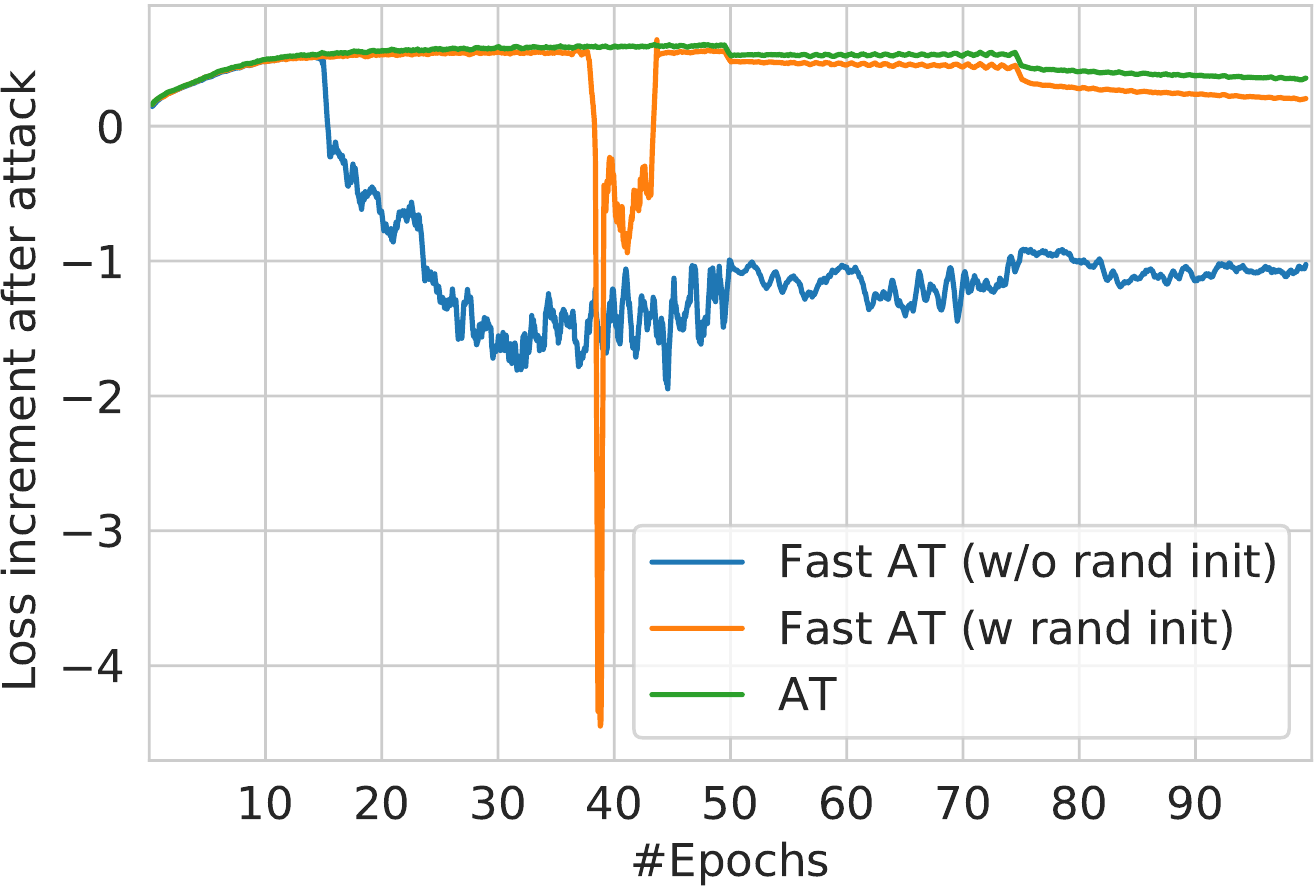}
  \caption{Loss increment after attack, i.e., $ L(f_{\btheta}(\xb + \bm{\delta}^*), y) - L(f_{\btheta}(\xb), y)$, along the training trajectory for different methods on training ResNet-18 on CIFAR-10 dataset.}
  \label{fig:loss_increment}
\end{figure}

In Figure \ref{fig:loss_increment}, we plot the loss increment $\Delta_L$ for three different training trajectories: Fast AT without random initialization, Fast AT with random initialization, as well as standard AT. We observe that with the random initialization, Fast AT's loss increment is quite close to standard AT (although it still can go wrong from time to time). 
However, without random initialization, the loss value after the attack is actually worse than before, suggesting the algorithm completed failed in solving \eqref{eq:inner_max}. Since Fast AT has only one step budget for the attack, this further implies the attack step size is too large to cause divergence in the gradient descent procedure. Yet on the other hand, due to the one step attack budget, the attack step size has to be chosen close to $\epsilon$ for better defense purposes\footnote{With a much smaller attack step size to $\epsilon$ and only one step attack budget, the generated adversarial examples during the training phase can never reach the magnitude of $\epsilon$. Therefore, when facing perturbations of the magnitude of $\epsilon$ during the testing phase, the model stands little chance defending against them.}. 
This dilemma explains the cause of failure for one-step AT without random initialization.

\subsection{Why Random Initialization Helps?}
Now let us talk about random initialization. It is well known from optimization theory \citep{boyd2004convex} that, for gradient descent-based algorithms, the maximum allowed step size (in order to guarantee convergence) is directly related to the smoothness of the optimization objective function. Specifically, the smoother the objective function is, the larger the gradient step size is allowed. Here we argue that random initialization works just as the randomized smoothing technique \citep{duchi2012randomized}, which makes the overall optimization objective more smooth via random perturbations of the optimization variable\footnote{Instead of using only the gradient at the original iterate, randomized smoothing proposes to randomly generate perturbed iterates and use their gradients for the optimization procedure.  More details about the randomized smoothing technique are provided in Appendix \ref{appendix:rs}.}. Note that this randomized smoothing is an optimization technique that is different from the Randomized Smoothing method in certified defenses \citep{cohen2019certified}, although the names are the same.  

To see why random initialization works as randomized smoothing in Fast AT, let us apply randomized smoothing to \eqref{eq:inner_max} and we have:
\begin{align}\label{eq:inner_max_smooth}
     \bm{\delta}^* = \argmax_{\bm{\delta}+u\bxi \in \cB_{\epsilon}(\bm{0})} \EE_{\bxi \sim U(-1, 1)} L(f_{\btheta}(\xb + \bm{\delta} + u\bxi), y),
\end{align}
where $\bxi$ is the perturbation vector for randomized smoothing, $u$ controls the smoothing effect, and $\bm{\delta}$ is the adversarial perturbation vector (initialized as zero). Suppose we have $u=\epsilon$ and solve \eqref{eq:inner_max_smooth} in a stochastic fashion (\emph{i.e.}, sample a random perturbation $\bxi$ instead of computing the expectation over $\bxi$), and using only one step gradient update, it reduces to the Fast AT formulation. 
This suggests that Fast AT can be viewed as performing stochastic single-step attacks on a randomized smoothed objective function which allows the use of larger step size. This explains why random initialization helps Fast AT in Figure \ref{fig:loss_increment}: as it makes the loss objective smoother, thus become easier to optimize with large step sizes and avoid possible divergence cases.

It is worth noting that \cite{andriushchenko2020understanding} also provided an explanation of random initialization: it reduces the magnitude of the perturbation and thus the network becomes more linear and fits better toward single-step attack. 
In fact, our argument is more general and can cover theirs, because if the loss function is approximately linear, then it will be very smooth, \emph{i.e.}, the second-order term in the Taylor expansion is small. And their observations that Fast AT using smaller attack step size can succeed without random initialization actually also validate our analysis above.

\subsection{Drawbacks of Random Initialization}\label{sec:problem_rand_init}
Although the random initialization effectively helps Fast AT avoid the catastrophic overfitting from happening in the most time, it still exposes several major weaknesses.

\noindent\textbf{Performance Stability}\,
Fast AT can still be highly unstable (\emph{i.e.}, catastrophic overfitting can still occur from time to time). This is also observed in \cite{li2020towards}. In Figure \ref{fig:loss_increment}, we also observe that Fast AT could still fail in solving the inner maximization problem (especially when using a drastically large attack step size). It can be imagined that with some bad luck, the training procedure of Fast AT could still fall apart even with random initialization.




Unfortunately, to get the best from Fast AT, it usually requires a larger attack step size. We run Fast AT on CIFAR-10 using ResNet-18 model \citep{he2016deep} for $10$ times\footnote{Here we exclude the additional acceleration techniques in \cite{Wong2020Fast} and apply standard piecewise learning rate decay as in \cite{madry2017towards, zhang2019theoretically}.}. For the best attack step size of $10/255$ (according to \citep{Wong2020Fast}), the best run achieves $46.30\%$ robust accuracy, however, the average is only $42.11\%$ since many runs actually failed.


\begin{table}[t!]
\setlength{\abovecaptionskip}{-4pt}
\small
\caption{Model robustness comparison among AT, Fast AT, TRADES and Fast TRADES, using ResNet-18 model on CIFAR-10 dataset.}
\vskip 0.1in
\label{table:fast_compare}
\centering
\begin{tabular}{lcc}
\toprule
\multicolumn{1}{l}{\bf Method}  &\multicolumn{1}{c}{\bf Nat (\%)} &\multicolumn{1}{c}{\bf Rob (\%)}\\
\midrule 
AT & 82.36 & 51.14\\
Fast AT & 84.79 & 46.30 \\
TRADES & 82.33 & 52.74\\
Fast TRADES & 83.39 & 46.98\\
\bottomrule
\end{tabular}
\\
\tiny{*\textbf{Nat}: accuracy evaluated on the clean test examples; \\ *\textbf{Rob}: accuracy evaluated on adversarial examples of the test set.}
\vspace{-3pt}
\end{table}

\noindent\textbf{Further Robustness Improvement}\, Fast AT uses standard adversarial training \citep{madry2017towards} as the baseline, and can obtain similar robustness performance. However, later work \citep{rice2020overfitting} shows that original adversarial training's performance is deteriorated by robust overfitting, while simply using early stopping can largely improve its robustness. \cite{zhang2019theoretically} further achieves even better model robustness that is much higher than what Fast AT obtains. From Table \ref{table:fast_compare}, we observe that there exists a $6\%$ robust accuracy gap on the CIFAR-10 dataset between Fast AT and TRADES. This indicates that Fast AT is still far from optimal, and there is still big room for further robustness improvement.



\section{Proposed Approaches}\label{sec:method}

\subsection{A Naive Try: Randomized Smoothing for TRADES}\label{sec:fast_trades}
In the previous section, we show that objective smoothness plays a key role in the success of single-step adversarial training. Note the TRADES \citep{zhang2019theoretically} method naturally promotes the objective smoothness in its training formula (by minimizing the output discrepancy of input examples within the perturbation ball). From this perspective, it should be more fit to single-step robust training than AT. 
Therefore we try to apply randomized smoothing to TRADES and see if this leads to a better robust training method.
Let us recall the inner maximization formulation for TRADES:
\begin{align}
     \max_{\bm{\delta} \in \cB_{\epsilon}(\zero)} \text{KL}\big(s(f_{\btheta}(\xb)), s(f_{\btheta}(\xb + \bm{\delta})) \big),
\end{align}
where $s(\cdot)$ denotes the softmax function. 
Similarly, we can further apply randomized smoothing technique on this objective and obtain:
\begin{align}
     \max_{\bm{\delta} + u\bxi \in \cB_{\epsilon}(\bm{0})} \EE_{\bxi \sim U(-1, 1)} \text{KL}\big(s(f_{\btheta}(\xb)), s(f_{\btheta}(\xb + \bm{\delta} + u\bxi))\big).
\end{align}



Then we can apply the same stochastic single step attack and $u=\epsilon$ for solving this problem, i.e., first do random initialization and then perform single-step projected gradient ascent on TRADES loss. We refer to this strategy as Fast TRADES.
We experimentally test Fast TRADES by training the ResNet-18 model on the CIFAR-10 dataset. From Table \ref{table:fast_compare}, we can see that Fast TRADES indeed achieves slightly better performance than Fast AT. Yet it still falls far behind the original TRADES method. 
This inspires us to study how to design a better strategy for more significant improvements.

Note that our best performing Fast TRADES model in Table \ref{table:fast_compare} is obtained with attack step size $6/255$ (in contrast to $10/255$ for Fast AT). According to our previous analysis in Section \ref{sec:why}, if we can make the loss objective even smoother, it is possible to utilize an even larger attack step size for better robust training performances.
However, unlike the general randomized smoothing setting, where we can simply use a larger value of $u$ for a smoother objective, in the adversarial setting, the random perturbation on the input vector is subject to the $\epsilon$-ball constraint. This means that simply using larger $u$ cannot bring us a smoother loss objective, instead, we need to find new ways for better smoothing effects.



\subsection{Backward Smoothing}
\begin{figure}
  \centering
    \includegraphics[width=0.4\textwidth]{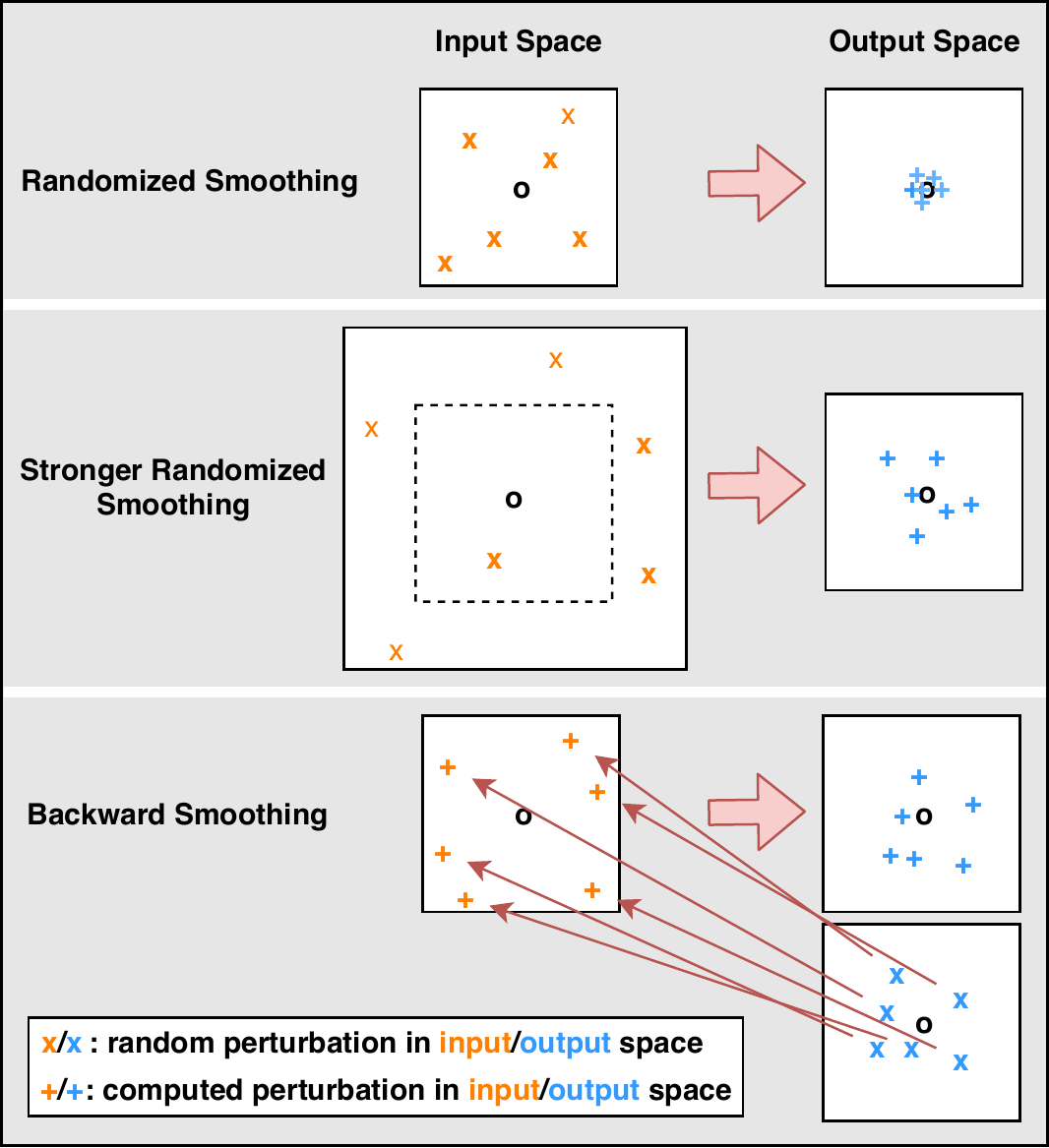}
    %
    \caption{A sketch of our proposed method.}
    \label{fig:output_smoothing}
\vskip -0.2in
\end{figure}

Now we introduce our proposed method to further boost the smoothing effect without violating the $\epsilon$-ball constraint. 
Let us denote the input domain $\xb \in \RR^d$ as the input space, and their corresponding neural network output $f_{\btheta}(\xb) \in \RR^c$ as the output space, where $c$ is the number of classes for the classifier. Note that if we have random samples in the input space, the corresponding output is actually quite close \citep{tashiro2020diversity} as in the first row of Figure \ref{fig:output_smoothing}.
Imagine that if we are allowed to use a larger $u$, the output space would be more diverse as in the second row of Figure \ref{fig:output_smoothing}.
This inspires us to generate the initialization point in a backward fashion. 
We first generate random points in the output space just as randomized smoothing does in the input space (see third row of Figure \ref{fig:output_smoothing}, lower right plot), \emph{i.e.}, $f_{\btheta} (\xb) + \gamma\bpsi$, where $\bpsi \sim U(-1,1)$ is the random variable
and $\gamma$ is a small number.
Then we find the corresponding input perturbation in a backward fashion and use it as our initialization. 
An illustrative sketch of our proposed method is provided in Figure \ref{fig:output_smoothing}.
In summary, we aim to find the input perturbation $\bxi$ such that: 
\begin{align}\label{eq:equal}
    f_{\btheta} (\xb + \bxi) = f_{\btheta} (\xb) + \gamma\bpsi.
\end{align}
In order to find the best $\bxi^*$ to satisfy \eqref{eq:equal}, we turn to solve the following problem:
\begin{align}\label{eq:output_smoothing_input}
    \bxi^* = \argmin_{\bxi \in \cB_{\epsilon}(\bm{0})} \text{KL}\big( s( f_{\btheta} (\xb ) + \gamma\bpsi ), s( f_{\btheta} (\xb  + \bxi) )\big).
\end{align}
Note that $\bxi$ is initialized as a zero vector.
For the sake of computational efficiency, we solve \eqref{eq:output_smoothing_input} using single-step PGD in practice.
Then, similar to \cite{Wong2020Fast}, we use single-step gradient update for the inner maximization problem:
\begin{align}\label{eq:output_smoothing}
     \bm{\delta}^* = \argmax_{\bm{\delta} + \bxi^* \in \cB_{\epsilon}(\bm{0})}  \text{KL}\big(s(f_{\btheta}(\xb)), s(f_{\btheta}(\xb + \bm{\delta} + \bxi^*))\big).
\end{align}
Finally, we update the neural network parameter $\btheta$ using stochastic gradients at $\xb + \bxi^* + \bm{\delta}^*$. A summary of our proposed algorithm is provided in Algorithm \ref{alg:output-smoothing}.

\begin{algorithm}[ht!]
	\caption{Backward Smoothing}
	\label{alg:output-smoothing}
	\begin{algorithmic}[1]
		\STATE \textbf{input:} The number of training iterations $T$, number of adversarial perturbation steps $K$, maximum perturbation strength $\epsilon$, training step size $\eta$, adversarial perturbation step size $\alpha $, regularization parameter $\beta > 0$;
		\STATE Random initialize model parameter $\btheta_0$
 		\FOR {$t = 1,\ldots, T$}
		      \STATE Sample mini-batch $\{\xb_i, y_i\}_{i=1}^m$ from training set
		      \STATE Obtain $\bxi^*$ by solving  \eqref{eq:output_smoothing_input} 
              \STATE Obtain $\bm{\delta}^*$ by solving \eqref{eq:output_smoothing} 
		      \STATE $\btheta_t = \btheta_{t-1} - \eta/m \cdot \sum_{i=1}^m  \nabla_{\btheta}\big[L(f_{\btheta}(\xb_i), y_i) + \beta \cdot \text{KL}\big(s(f_{\btheta}(\xb_i)), s(f_{\btheta}(\xb_i + \bxi^* + \bm{\delta}^*))\big)  \big]$
		\ENDFOR     
 	\end{algorithmic}
\end{algorithm}

\begin{figure}[t!]
  \centering
  \includegraphics[width=.4\textwidth]{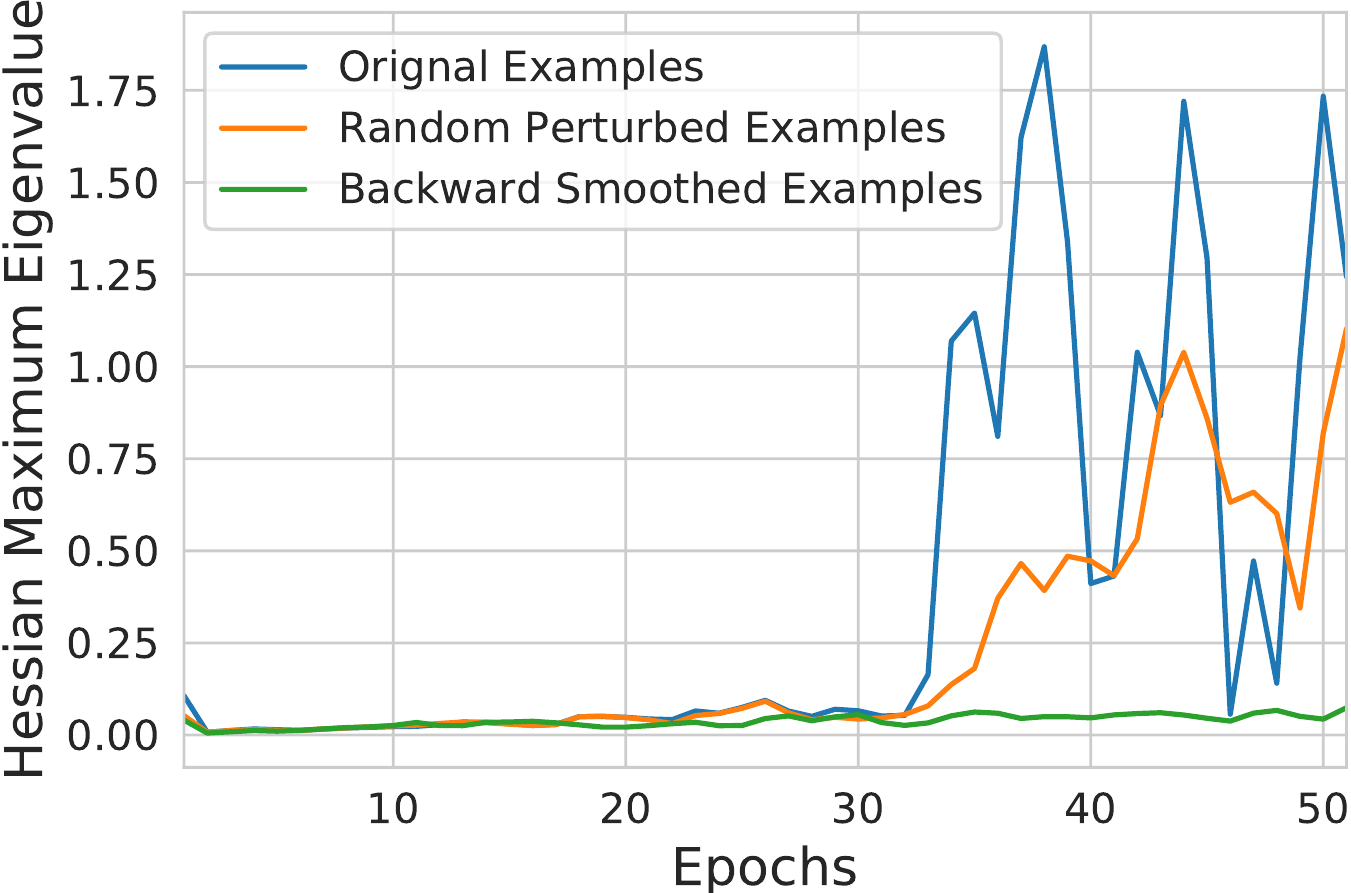}
  \caption{Hessian maximum eigenvalue comparison against training epochs.}
  \label{fig:smooth}
  \vspace{-3mm}
\end{figure}
Figure \ref{fig:smooth} shows the maximum eigenvalue of Hessian of the loss function at the original examples, randomly perturbed examples, and backward smoothed examples  along the training trajectory until Fast TRADES obtains its best robustness (the 51st epoch). We observe that during the model training process, the randomly perturbed examples have overall smaller Hessian maximum eigenvalue\footnote{The smaller Hessian maximum eigenvalue, the smoother the loss function is.} than that of original examples. This suggests that random smoothing indeed makes the loss function smoother. Moreover, the Hessian maximum eigenvalue under backward smoothing is much smaller than that under random smoothing, showing the insufficiency of the random smoothing techniques and the advantages of our proposed backward smoothing method.

\section{Experiments}\label{sec:exp}
In this section, we empirically evaluate the performance of our proposed method.
We first compare our proposed method with other robust training baselines on CIFAR-10, CIFAR100 \citep{krizhevsky2009learning} and Tiny ImageNet \citep{deng2009imagenet}\footnote{We do not test on ImageNet dataset mainly due to that TRADES does not perform well on ImageNet as mentioned in \cite{qin2019adversarial}.} datasets.
We also provide multiple ablation studies as well as robustness evaluation with state-of-the-art adversarial attack methods to validate that our proposed method provides effective robustness improvement. 

\subsection{Experimental Setting}
Following previous work on robust training \citep{madry2017towards,zhang2019theoretically,Wong2020Fast}, we set $\epsilon = 0.031$ for all three datasets. In terms of model architecture, we adopt standard ResNet-18 model \citep{he2016deep} for both CIFAR-10 and CIFAR-100 datasets, and ResNet-50 model for Tiny ImageNet. We follow the standard piecewise learning rate decay schedule as used in \cite{madry2017towards,zhang2019theoretically} and set decaying point at $50$-th and $75$-th epochs. The starting learning rate for all methods is set to $0.1$, the same as previous work \citep{madry2017towards,zhang2019theoretically}. 
For all methods, we tune the models for their best robustness performances for a fair comparison. 
For Adversarial Training and TRADES methods, we adopt a $10$-step iterative PGD attack with a step size of $2/255$ for both.
For our proposed method, we set the backward smoothing parameter $\gamma = 1$ and step size as $8/255$.
For other fast training methods, we use a step size of $10/255$ for Fast AT/GradAlign, $6/255$ for 2-step Fast AT, $6/255$ for Fast TRADES and $5/255$ for 2-step Fast TRADES. 
For robust accuracy evaluation, we typically adopt a $100$-step PGD attack with the step size of $2/255$. To ensure the validity of the model robustness improvement is not because of the obfuscated gradient \citep{athalye2018obfuscated}, we further test our method with current state-of-the-art attacks \citep{croce2020reliable,chen2020rays}. All the experiments are conducted on RTX2080Ti GPU servers.

\subsection{Performance Comparison with Robust Training Baselines}
We compare the adversarial robustness of Backward Smoothing against standard Adversarial Training \citep{madry2017towards}, TRADES \citep{zhang2019theoretically}, as well as fast training methods such as Fast AT \citep{Wong2020Fast} and our naive baseline Fast TRADES. We also compare with recently proposed Fast AT+ \citep{li2020towards}\footnote{Since \cite{li2020towards} does not have code released yet, we only compare with theirs in the same setting (combined with acceleration techniques) using reported numbers.} and GradAlign \citep{andriushchenko2020understanding}\footnote{We only compare with \cite{andriushchenko2020understanding} in Tables \ref{table:cifar10}, \ref{table:cifar100}, \ref{table:cifar10_tricks} as its double backpropagation formulation requires much larger memory usage.}.
Since our proposed backward smoothing initialization utilizes an extra step of gradient back-propagation, we also compare with Fast AT, Fast TRADES using 2-step attack for a fair comparison. 

\begin{table}[t!]
\setlength{\belowcaptionskip}{-2pt}
\setlength{\tabcolsep}{0.3em}
\begin{minipage}{.48\textwidth}
\centering
\caption{Performance comparison on CIFAR-10 using ResNet-18 model.}
\label{table:cifar10}
\begin{small}
\begin{tabular}{lccc}
\toprule 
\multicolumn{1}{l}{\bf Method}  &\multicolumn{1}{c}{\bf Nat (\%)} &\multicolumn{1}{c}{\bf Rob (\%)}
&\multicolumn{1}{c}{\bf Time (m)}\\
\midrule 
AT  & 82.36 &
51.14 & 430\\
Fast AT & \textbf{84.79} & 46.30 & \textbf{82}\\
Fast AT (2-step) &  83.21 &  49.91 &  {127}\\
Fast AT (GradAlign) & 84.37 & 46.99 & 402 \\
TRADES  & 82.33 & \textbf{52.74} & 482\\
Fast TRADES  & 83.39 & 46.98 & 126\\
Fast TRADES (2-step)  & 83.51 & 48.78 & 164\\
\textit{Backward Smoothing} & 82.38
 & 52.50 & 164 \\
\bottomrule 
\end{tabular}
\end{small}
\end{minipage} 
\end{table}

\begin{figure}[t!]
  \begin{center}
    \includegraphics[width=0.4\textwidth]{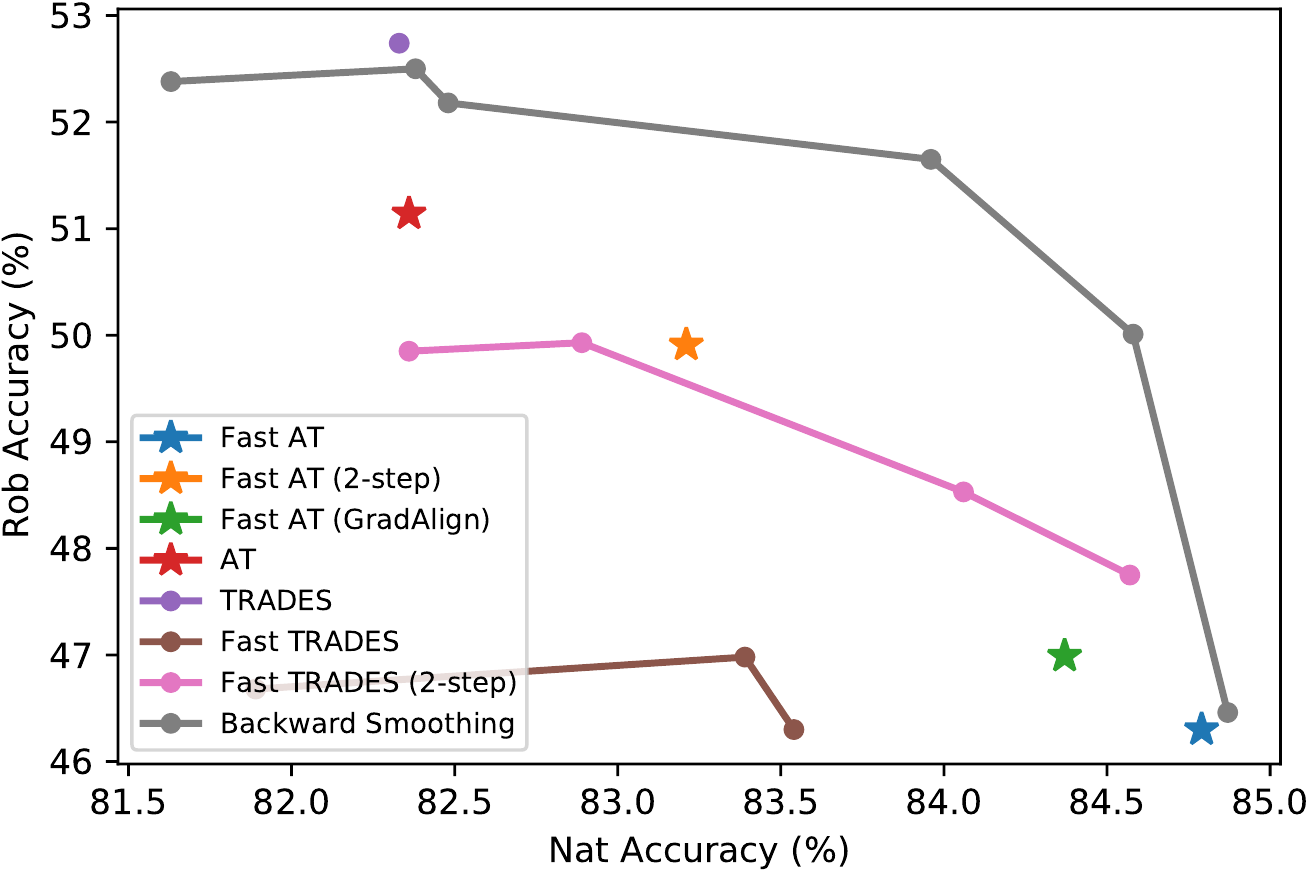}
    \caption{Backward Smoothing's performance gain is not due to robustness-accuracy trade-off.
    }
    \label{fig:tradeoff}
  \end{center}
  \vspace{-5mm}
\end{figure}

\begin{figure}[t!]
  \begin{center}
    \includegraphics[width=0.4\textwidth]{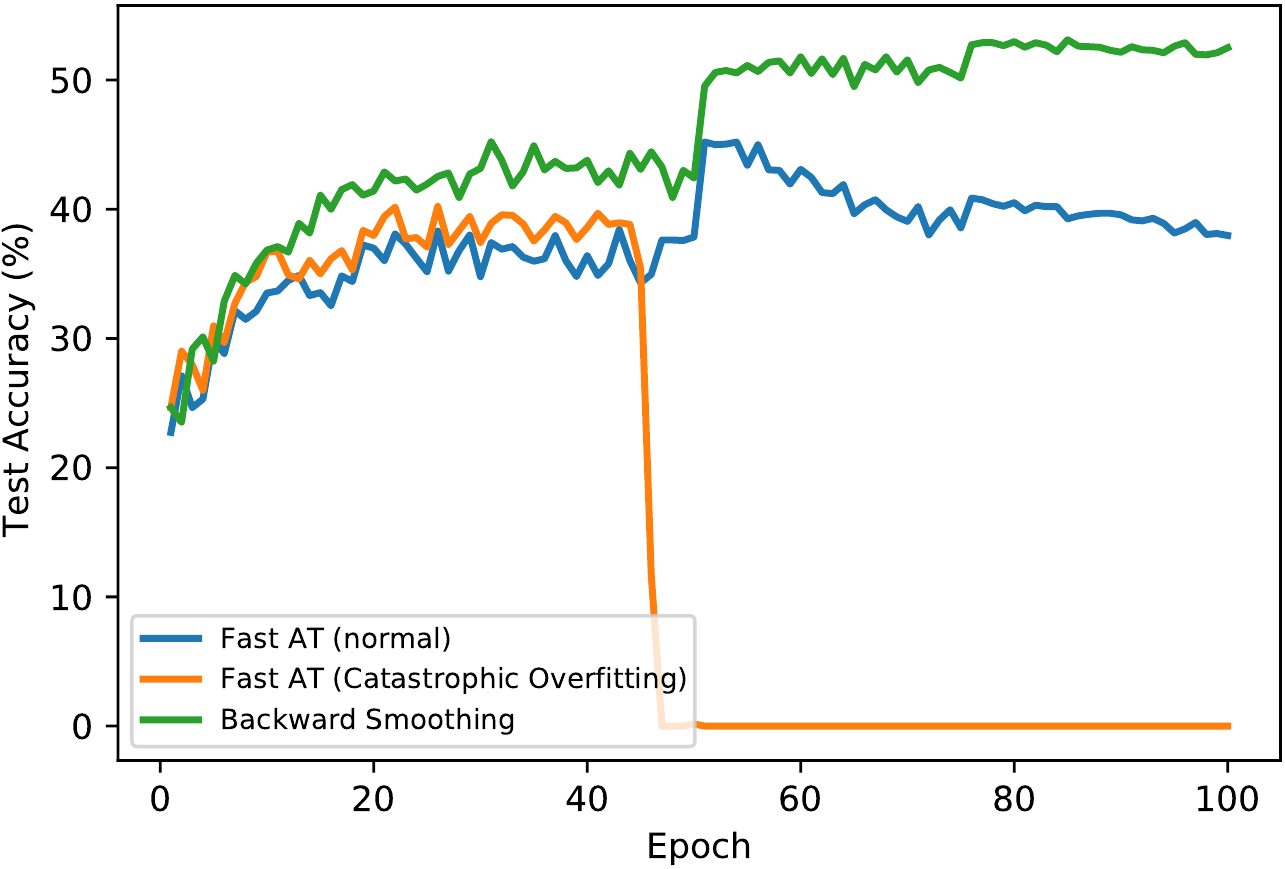}
    \caption{Backward Smoothing does not suffer from the catastrophic overfitting phenomenon.
    }
    \label{fig:overfitting}
  \end{center}
  \vspace{-5mm}
\end{figure}

Table \ref{table:cifar10} shows the performance comparison on the CIFAR-10 dataset using ResNet-18 model. Our Backward Smoothing method achieves high robust accuracy that is almost as good as state-of-the-art methods such as TRADES, while consuming much less ($\sim$3x) training time. Compared with Fast AT, Backward Smoothing typically costs twice the training time, yet achieving significantly higher model robustness. Notice that  the GradAlign method indeed slightly improves upon Fast AT, but it also costs much more training time due to its double backpropagation formulation, making it less competitive to our Backward Smoothing method. Our method also achieves a large performance gain against Fast TRADES. Note that even compared with Fast TRADES using 2-step attack and Fast AT using 2-step attack, which costs about the same training time as ours, our method still achieves a large improvement. 

Note that \citet{zhang2019theoretically} has shown that there exists a  robustness-accuracy trade-off in robust training. In order to make sure that our proposed method's performance gain is not due to this robustness-accuracy trade-off, we further test with different choices of robust regularization parameter $\beta$ and plot the robust accuracy against natural accuracy plot in Figure \ref{fig:tradeoff}. Note that for AT and Fast AT or GradAlign method, their formulations do not contain any tunable parameters for this robustness-accuracy trade-off, therefore, we only plot the single point for them. From Figure \ref{fig:tradeoff}, we can observe that the Backward Smoothing method indeed largely outperforms the other fast training baselines (achieve better robustness under roughly the same natural accuracy), and is not due to balancing the robustness-accuracy trade-off.
In Figure \ref{fig:overfitting}, we further verify whether Backward Smoothing still suffers from the catastrophic overfitting phenomenon. Specifically, we plot the test accuracy against the training epochs for Fast AT (normal), Fast AT (with catastrophic overfitting) and Backward Smoothing. As can be seen from Figure \ref{fig:overfitting}, compared to Fast AT, the Backward Smoothing method actually helps mitigate overfitting at the later stage of training.

\begin{table}[t!]
\setlength{\tabcolsep}{0.3em}
\centering
\caption{Performance comparison on CIFAR-100 using ResNet-18 model.}
\label{table:cifar100}
\begin{small}
\begin{tabular}{lccc}
\toprule 
\multicolumn{1}{l}{\bf Method}  &\multicolumn{1}{c}{\bf Nat (\%)} &\multicolumn{1}{c}{\bf Rob (\%)}
&\multicolumn{1}{c}{\bf Time (m)}\\
\midrule 
AT & 55.22 & 28.53 & 428\\
Fast AT & \textbf{60.35}  & 24.64 & \textbf{83}\\
Fast AT (2-step) &  56.00  &  27.84 &  128\\
Fast AT (GradAlign) & 58.38 & 26.26 & 402 \\
TRADES  & 56.99 &  29.41 & 480\\
Fast TRADES  & 60.26 & 21.33 & 126\\
Fast TRADES (2-step) &  58.81 & 25.47 &
165\\
\textit{Backward Smoothing} & 56.96 & \textbf{30.50} & 164 \\
\bottomrule 
\end{tabular}
\end{small}
\vspace{-0.1in}
\end{table}

\begin{table}[t!]
\setlength{\tabcolsep}{0.3em}
\setlength{\belowcaptionskip}{-5pt}
\caption{Performance comparison on Tiny ImageNet dataset using ResNet-50 model.}
\label{table:tiny}
\centering
\small
\begin{tabular}{lccc}
\toprule
\multicolumn{1}{l}{\bf Method}  &\multicolumn{1}{c}{\bf Nat (\%)} &\multicolumn{1}{c}{\bf Rob (\%)}
&\multicolumn{1}{c}{\bf Time (m)}\\
\midrule 
AT & 44.50 & 
21.34 & 
2666\\
Fast AT &  \textbf{49.58} & 
18.56 & 
\textbf{575}\\
 Fast AT (2-step) &  45.74 & 
 20.94 &  817\\
TRADES  & 47.02 & 
21.04 & 
2928\\
Fast TRADES & 50.36 & 
17.22 & 
805\\
Fast TRADES (2-step)  & 46.92 & 19.26 & 
1045\\
\textit{Backward Smoothing} & 46.68 &
\textbf{22.32} & 
1035\\
\bottomrule
\end{tabular}
\vspace{-0.1in}
\end{table}

Table \ref{table:cifar100} shows the performance comparison on CIFAR-100 using ResNet-18 model. We can observe patterns similar to the CIFAR-10 experiments. Backward Smoothing achieves slightly higher robustness compared with TRADES, while costing much less training time. 
Compared with Fast TRADES using 2-step attack and Fast AT using 2-step attack, our method also achieves a large robustness improvement with roughly the same training cost.
Table \ref{table:tiny} shows that on Tiny ImageNet using the ResNet-50 model, Backward Smoothing also achieves significant robustness improvement over other single-step robust training methods.

\subsection{Evaluation with State-of-the-art Attacks}
To ensure that Backward Smoothing does not cause obfuscated gradient problem \citep{athalye2018obfuscated} or presents a false sense of security, we further evaluate our method using state-of-the-art attacks, by considering two evaluation methods: $(i)$ AutoAttack \citep{croce2020reliable}, which is an ensemble of four diverse (white-box and black-box) attacks (APGD-CE, APGD-DLR, FAB \citep{croce2020minimally} and Square Attack \citep{andriushchenko2020square}) to reliably evaluate robustness; $(ii)$ RayS attack \citep{chen2020rays}, which only requires the prediction labels of the target model (completely gradient-free) and is able to detect falsely robust models. It also measures another robustness metric, average decision boundary distance (ADBD), defined as examples' average distance to their closest decision boundary. ADBD reflects the overall model robustness beyond $\epsilon$ constraint. Both evaluations provide online robustness leaderboards for public comparison with other models.

\begin{table}[h]
\setlength{\belowcaptionskip}{-5pt}
\setlength{\tabcolsep}{0.3em}
\caption{Performance comparison with SOTA robust models on CIFAR-10 evaluated by AutoAttack and RayS.}
\label{table:leaderboard}
\centering
\small
\begin{tabular}{lccc}
\toprule
\multicolumn{1}{l}{\bf Method}  &\multicolumn{1}{c}{\bf AutoAttack}&\multicolumn{2}{c}{\bf RayS}\\
\multicolumn{1}{l}{\bf Metric}  &\multicolumn{1}{c}{\bf Rob (\%)}&\multicolumn{1}{c}{\bf Rob (\%)} &\multicolumn{1}{c}{\bf ADBD}\\
\midrule 
AT (original, no early-stop) & 44.04 &  50.70 & 0.0344\\
 AT  &  49.10 &  54.00 &  0.0377\\
Fast AT & 43.21 & 50.10 & 0.0334\\
TRADES & \textbf{53.08} & \textbf{57.30} & \textbf{0.0403}\\
 Fast TRADES &  43.84 &  52.05 &  0.0348 \\
 Fast TRADES (2-step) &   48.20 &  54.43 &  0.0383\\
\textit{Backward Smoothing}  & 51.13 & 55.08 & \textbf{0.0403}\\
\bottomrule
\end{tabular}
\vspace{-0.1in}
\end{table}

We train our method with WideResNet-34-10 model \citep{zagoruyko2016wide} and evaluate via AutoAttack and RayS. Table \ref{table:leaderboard} shows that under state-of-the-art attacks, Backward Smoothing still holds high robustness comparable to TRADES. Specifically, in terms of robust accuracy, Backward Smoothing is only $2\%$ behind TRADES, while significantly higher than AT \citep{madry2017towards} and Fast AT \citep{Wong2020Fast}. In terms of ADBD metric, Backward Smoothing achieves the same level of overall model robustness as TRADES, much higher than the other two methods.
Note that the gap between Backward Smoothing and TRADES is larger than that in Table \ref{table:cifar10}. We want to emphasize that this is not mainly due to the stronger attacks but the fact that we are using larger model architectures. Intuitively speaking, larger models have larger capacities and may need stronger attacks to reach some dark spot in the area. 

\begin{table}[h]
\caption{Performance comparison on CIFAR-10 using ResNet-18 model combined with cyclic learning rate and mix-precision training.}
\setlength{\tabcolsep}{0.3em}
\label{table:cifar10_tricks}
\centering
\small
\begin{tabular}{lccc}
\toprule
\multicolumn{1}{l}{\bf Method}  &\multicolumn{1}{c}{\bf Nat (\%)} &\multicolumn{1}{c}{\bf Rob (\%)}
&\multicolumn{1}{c}{\bf Time (m)}\\
\midrule
AT  &  81.48 & 50.32  & 62\\
Fast AT & 83.26 & 45.30 & \textbf{12}\\
Fast AT+ & 83.54 & 48.43 & 28\\
Fast AT (GradAlign) & 81.80 & 46.90 & 54 \\
TRADES  & 79.64 & \textbf{50.86} & 88\\
Fast TRADES  & \textbf{84.40} & 45.96 & 18\\
Fast TRADES (2-step)  & 81.37  & 47.56 &  24 \\
\textit{Backward Smoothing} & 78.76 & 50.58 & 24 \\
\bottomrule
\end{tabular}
\vspace{-0.15in}
\end{table}

\subsection{Combining with Other Acceleration Techniques}

Aside from random initialization, \cite{Wong2020Fast} also adopts two additional acceleration techniques to further improve training efficiency with a minor sacrifice on robustness performance: cyclic learning rate decay schedule \citep{smith2017cyclical} and mix-precision training \citep{micikevicius2018mixed}. We show that such strategies are also applicable to Backward Smoothing. 
Table \ref{table:cifar10_tricks} provides the results when these acceleration techniques are applied. We can observe that both work universally well for all methods, significantly reducing training time (in comparison  with Table \ref{table:cifar10}). Yet it does not alter the conclusions that Backward Smoothing achieves similar robustness to TRADES with much less training time. Also when compared with the recent proposed Fast AT+ method, Backward Smoothing achieves higher robustness and training efficiency. Note that the idea of the Fast AT+ method is orthogonal to ours and can be adopt to ours for further reduction on training time.

\section{Conclusions}\label{sec:con}
In this paper, we analyze the reason why single-step robust training without random initialization would fail and propose a new understanding towards Fast Adversarial Training by viewing random initialization as performing randomized smoothing for the inner maximization problem. Following this new perspective, we further propose a new initialization strategy, Backward Smoothing. The resulting method avoids the catastrophic overfitting problem and improves the robustness-efficiency trade-off over previous single-step robust training methods.


\bibliography{adv}

\appendix
\newpage
\onecolumn

\section{Randomized Smoothing}\label{appendix:rs}
Randomized smoothing technique \citep{duchi2012randomized} was originally proposed for solving convex non-smooth optimization problems.
It is based on the observations that random perturbation of the optimization variable can be used to transform the loss into a smoother one.
Instead of using only $L(\xb)$ and $\nabla L(\xb)$ to solve $$\min L(\xb),$$
randomized smoothing turns to solve the following objective function, which utilizes more global information from neighboring areas: 
\begin{align}\label{eq:smoothing}
    \min \EE_{\bxi \sim U(-1, 1)} L(\xb + u\bxi),
\end{align}
where $\bxi$ is a random variable, and $u$ is a small number.  
\cite{duchi2012randomized} showed that randomized smoothing makes the loss in \eqref{eq:smoothing} smoother than before. Hence, even if the original loss $L$ is non-smooth, it can still be solved by stochastic gradient descent with provable guarantees.



\section{Additional Experiments}
In this section, we conduct additional experiments to provide a comprehensive view to the Backward Smoothing method. Note that for all the experiments in the paper, we tune the hyperparameters for achieving the best robustness performances of each method. Specifically, we tune the attack step size from $[1/255, 2/255, ..., 10/255]$, vary the learning rate from $[0.001, 0.01, 0.1]$, $\gamma$ from $[0.1, 0.5, 1.0, 2.0, 5.0]$ and $\beta$ from $[2,4,...,10]$.

\subsection{Stability and Sensitivity}
In this subsection, we also study the stability and sensitivity of our proposed Backward Smoothing method. Due to the space limit, we only present two sets of experiments here. More ablation studies can be found in the Appendix.

\paragraph{Training Stability} We first take a look into the training stability. In Section \ref{sec:why} we have shown that Fast AT can still be highly non-stable in spite of its decent robustness performances. 
Figure \ref{fig:stability} shows that Backward Smoothing is much more stable than Fast AT with much smaller variances.  Compared with Fast TRADES, Backward Smoothing has achieved similar variance while obtaining much higher average model robustness. This demonstrates the superiority of our Backward Smoothing method on training stability.

\paragraph{Sensitivity of Attack Step Size}
We also take a look at the sensitivity of our Backward Smoothing method with various attack step sizes. From Table \ref{table:step}, we can observe that unlike Fast AT, which typically enjoys better robustness with larger step size (until it is too large and failed in training), Backward Smoothing achieves similar robustness with a slightly smaller step size, while the best performance is obtained with step size $8/255$. This suggests that we do not need to pursue overly-large step size for better robustness as in Fast AT. This also helps avoid the stability issue in Fast AT.
 
\vspace{0.1in}
\begin{minipage}{\textwidth}
\begin{minipage}{.45\textwidth}
\begin{center}
    \includegraphics[width=0.8\textwidth]{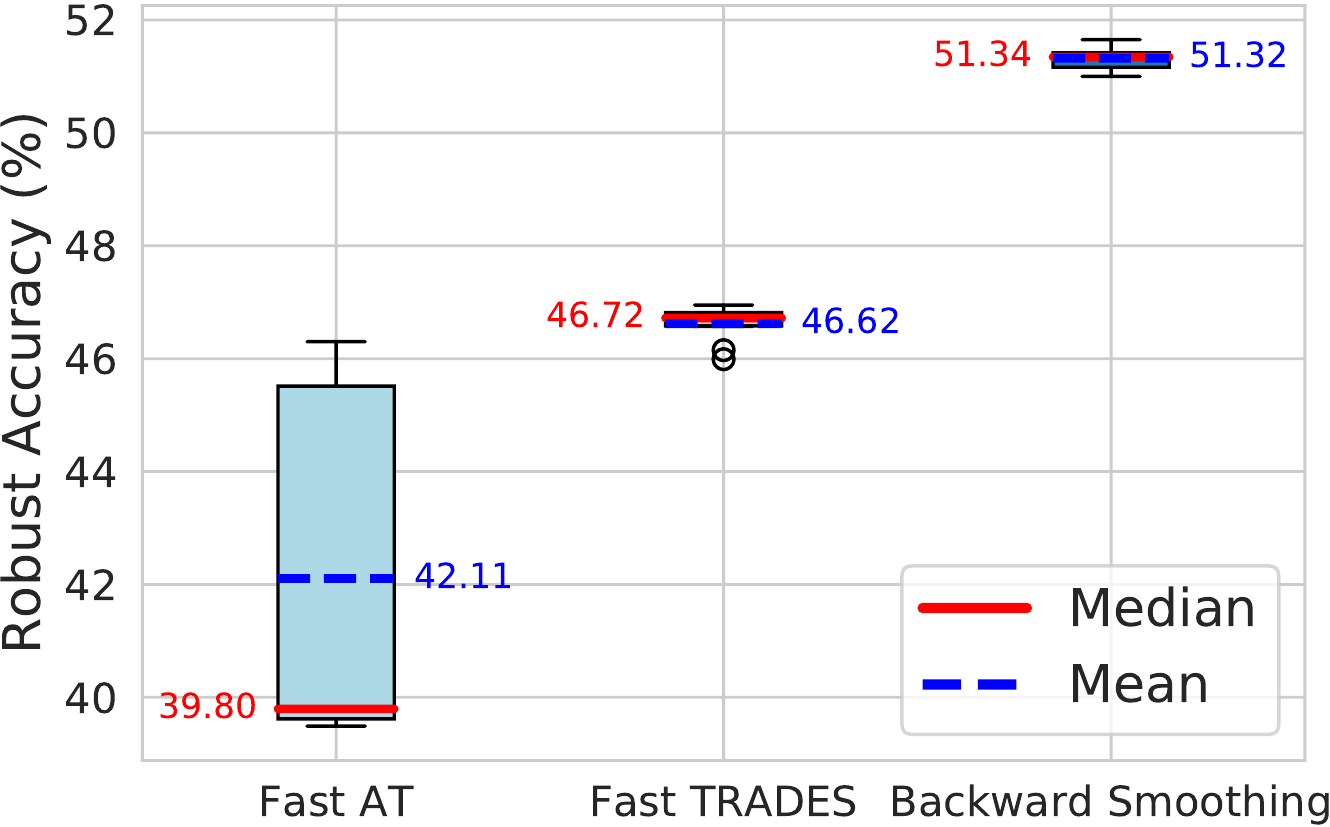}
    \captionof{figure}{Robustness stability of different fast robust training methods.
    }
    \label{fig:stability}
  \end{center}
\end{minipage}
\hfill
\begin{minipage}{.45\textwidth}
\captionof{table}{Sensitivity analysis of the attack step size on the CIFAR-10 and CIFAR-100 datasets using ResNet-18 model.}
\label{table:step}
\centering
\begin{small}
\setlength{\tabcolsep}{0.3em}
\begin{tabular}{ccccc}
\toprule
\multicolumn{1}{c}{\bf Dataset}  &\multicolumn{2}{c}{\bf CIFAR-10} &\multicolumn{2}{c}{\bf CIFAR-100}\\
\multicolumn{1}{c}{\bf Step Size}  &\multicolumn{1}{c}{\bf Nat (\%)} &\multicolumn{1}{c}{\bf Rob (\%)}
&\multicolumn{1}{c}{\bf Nat (\%)} &\multicolumn{1}{c}{\bf Rob (\%)}\\
\midrule 
6/255 & 81.38 & 52.38 & 56.83 & 29.78\\
7/255 & 81.96 & 52.40 & 56.61 & 29.82\\
8/255 & 82.38 & \textbf{52.50} & 56.96 & \textbf{30.50}\\
9/255 & 82.47 & 52.16 & 56.45 & 29.35\\
10/255 & 81.71 & 52.04 & 60.85 & 24.21\\
11/255 & 67.43 & 42.45 & 40.40 & 20.92\\
12/255 & 65.56 & 41.12 & 37.90 & 18.83\\
\bottomrule
\end{tabular}
\end{small}
\end{minipage}
\end{minipage}



\subsection{Does Backward Smoothing alone works?}
To further understand the role of Backward Smoothing in robust training, we conduct experiments on using Backward Smoothing alone, i.e., only use Backward Smoothing initialization but do not perform gradient-based attack at all. Table \ref{table:smoothing_only} and Table \ref{table:smoothing_only_cifar100} show the experimental results. We can observe that Backward Smoothing as an initialization itself only provides a limited level of robustness (not as good as a single-step attack). This is reasonable since the loss for Backward Smoothing does not directly promote adversarial attacks. Therefore it only serves as an initialization to help single-step attacks better solve the inner maximization problems.

\begin{table}[ht!]
\begin{minipage}{.48\textwidth}
\caption{Performance of using Backward Smoothing alone on CIFAR-10 dataset using ResNet-18 model.}
\label{table:smoothing_only}
\centering
\begin{small}
\begin{tabular}{lcc}
\multicolumn{1}{c}{\bf Method}  &\multicolumn{1}{c}{\bf Nat (\%)} &\multicolumn{1}{c}{\bf Rob (\%)}\\
\hline 
Fast AT & 84.79 & 46.30 \\
Fast TRADES  & 84.80 & 46.25 \\
Backward Smoothing Alone & 69.87 & 39.26  \\
\end{tabular}
\end{small}
\end{minipage}
\hfill
\begin{minipage}{.48\textwidth}
\caption{Performance of using Backward Smoothing alone on CIFAR-100 dataset using ResNet-18 model.}
\label{table:smoothing_only_cifar100}
\centering
\begin{small}
\begin{tabular}{lcc}
\multicolumn{1}{c}{\bf Method}  &\multicolumn{1}{c}{\bf Nat (\%)} &\multicolumn{1}{c}{\bf Rob (\%)}\\
\hline 
Fast AT & 60.35 & 24.64 \\
Fast TRADES  & 60.22 & 19.40 \\
Backward Smoothing Alone  & 43.47 &
18.51\\
\end{tabular}
\end{small}
\end{minipage}
\end{table}

\subsection{More Experiments for Backward Smoothing using Multiple Random Points}

We also conducted extra experiments using multiple random points for the Backward Smoothing method. As can be seen from Table \ref{table:num_rand_points}, a single random point already leads to similar performance as multiple random points but saves more time. Note that our target is to improve the efficiency of adversarial training, therefore, we only use a single random point for randomized smoothing in our proposed method.

\begin{table}[ht!]
\caption{Sensitivity analysis on the number of random points used in Backward Smoothing on the CIFAR-10 dataset using ResNet-18 model.}
\label{table:num_rand_points}
\centering
\begin{small}
\begin{tabular}{ccc}
\toprule
\multicolumn{1}{c}{\bf 
\# RandPoints} & \bf Rob (\%) & \bf Time (m) \\
\midrule 
1 & 52.50 & 	164\\
2 & 	52.67 & 	204\\
5 & 	52.70 & 	316\\
10 & 	52.73 & 	510\\
\bottomrule
\end{tabular}
\end{small}
\end{table}

\subsection{Ablation Studies on $\gamma$ and $\epsilon$}
We also perform a set of ablation studies to provide a more in-depth analysis on Backward Smoothing. 

\paragraph{Effect of $\gamma$}
We analyze the effect of $\gamma$ in Backward Smoothing by fixing $\beta$ and the attack step size. Table \ref{table:gamma} summarizes the results. In general, $\gamma$ does not have a significant effect on the final model robustness; however, using too large or too small $\gamma$ would lead to slightly worse robustness. Empirically, $\gamma=1$ achieves the best performance on both datasets.
 
\begin{table}[t!]
\setlength{\belowcaptionskip}{-2pt}
\setlength{\tabcolsep}{0.3em}
\begin{minipage}{.48\textwidth}
\caption{Sensitivity analysis of $\gamma$ on the CIFAR-10 and CIFAR-100 datasets using ResNet-18 model.}
\label{table:gamma}
\centering
\begin{small}
\begin{tabular}{ccccc}
\toprule
\multicolumn{1}{c}{\bf Dataset}  &\multicolumn{2}{c}{\bf CIFAR-10} &\multicolumn{2}{c}{\bf CIFAR-100}\\
\multicolumn{1}{c}{\bf $\gamma$}  &\multicolumn{1}{c}{\bf Nat (\%)} &\multicolumn{1}{c}{\bf Rob (\%)}
&\multicolumn{1}{c}{\bf Nat (\%)} &\multicolumn{1}{c}{\bf Rob (\%)}\\
\midrule
0.1  & 82.43 & 52.13 & 56.62 & 29.34\\
0.5 & 82.53 & 52.34 & 56.95  &  29.85\\
1.0 & 82.38 & \textbf{52.50} & 56.96 & \textbf{30.50}\\
2.0 & 82.29 & 52.42 & 56.16 & 29.88 \\
5.0 & 81.50 & 52.32 & 56.10 & 429.83\\
\bottomrule
\end{tabular}
\end{small}
\end{minipage} 
\hfill
 \begin{minipage}{.48\textwidth}
 \caption{Sensitivity analysis of $\beta$ on CIFAR-10 and CIFAR-100 datasets using ResNet-18 model.}
\label{table:beta}
\centering
\begin{small}
\begin{tabular}{ccccc}
\toprule
\multicolumn{1}{c}{\bf Dataset}  &\multicolumn{2}{c}{\bf CIFAR-10} &\multicolumn{2}{c}{\bf CIFAR-100}\\
\multicolumn{1}{c}{\bf $\beta$}  &\multicolumn{1}{c}{\bf Nat (\%)} &\multicolumn{1}{c}{\bf Rob (\%)}
&\multicolumn{1}{c}{\bf Nat (\%)} &\multicolumn{1}{c}{\bf Rob (\%)}\\
\midrule 
2.0 & 84.87 & 46.46 & 62.22 & 24.83\\
4.0 & 84.58 & 50.01 & 59.03 & 27.58\\
6.0 & 83.96 & 51.65 & 57.46 & 28.66\\
8.0 & 82.48 & 51.88 & 57.51 & 29.38\\
10.0 & 82.38 & \textbf{52.50} & 56.96 & \textbf{30.50}\\
12.0 & 81.63 & 52.38 & 56.46 & 29.95\\
\bottomrule
\end{tabular}
\end{small}
 \end{minipage} 
\end{table}

\paragraph{The Effect of $\beta$}
We conduct the ablation studies to figure out the effect of $\beta$ in the Backward Smoothing method by fixing $\gamma$ and the attack step size. Table \ref{table:beta} shows the experimental results. Similar to what $\beta$ does in TRADES \citep{zhang2019theoretically}, here in Backward Smoothing, $\beta$ still controls the trade-off between natural accuracy and robust accuracy. We observe that with a larger $\beta$, natural accuracy keeps decreasing and the best robustness is obtained with $\beta=10.0$.

\subsection{Experiments on different perturbation strength}
We also conducted experiments to compare the performance in other $\epsilon$ settings. Specifically, we compare the $\epsilon=4/255$ case and $\epsilon=12/255$ case in Table \ref{table:cifar10_eps4} and \ref{table:cifar10_eps12}. Both tables again show the advantages of our Backward Smoothing algorithm over other baselines.

\begin{table}[ht!]
\vspace{-3mm}
\centering
\caption{Performance comparison on CIFAR-10 using ResNet-18 model ($\epsilon = 4/255$).}
\label{table:cifar10_eps4}
\begin{small}
\begin{tabular}{lccc}
\toprule 
\multicolumn{1}{l}{\bf Method}  &\multicolumn{1}{c}{\bf Nat (\%)} &\multicolumn{1}{c}{\bf Rob (\%)}
&\multicolumn{1}{c}{\bf Time (m)}\\
\midrule 
AT  & 88.43 & 68.85 & 428\\
Fast AT & {89.40} & 65.80 & \textbf{90}\\
Fast AT (2-step) &  \textbf{89.50} & 66.89 & 129\\
Fast AT (GradAlign) & 89.15 & 65.78 & 401 \\
TRADES  & 88.35 & \textbf{70.05} & 478\\
Fast TRADES  & 89.20 & 66.71 & 136\\
Fast TRADES (2-step)  & 88.73 & 67.86 & 174\\
\textit{Backward Smoothing} & 87.22 & 69.67 & 165 \\
\bottomrule 
\end{tabular}
\end{small}
\end{table}

\begin{table}[ht!]
\centering
\caption{Performance comparison on CIFAR-10 using ResNet-18 model ($\epsilon = 12/255$).}
\label{table:cifar10_eps12}
\begin{small}
\begin{tabular}{lccc}
\toprule 
\multicolumn{1}{l}{\bf Method}  &\multicolumn{1}{c}{\bf Nat (\%)} &\multicolumn{1}{c}{\bf Rob (\%)}
&\multicolumn{1}{c}{\bf Time (m)}\\
\midrule 
AT  & 72.92 & \textbf{39.56} & 433\\
Fast AT & {64.06} & 26.14 & \textbf{90}\\
Fast AT (2-step) &  77.95 & 33.68 &  {127}\\
Fast AT (GradAlign) & 76.02 & 33.03 & 400 \\
TRADES  & \textbf{76.07} & {39.11} & 475\\
Fast TRADES  & 64.12 & 25.93 & 137\\
Fast TRADES (2-step)  & 75.98 & 29.91 & 174\\
\textit{Backward Smoothing} & 71.90 & 35.22 & 166\\
\bottomrule 
\end{tabular}
\end{small}
\end{table}

\subsection{PGD based Backward Smoothing}
We also wonder whether Backward Smoothing is compatible with Adversarial Training, \emph{i.e.}, can we use a similar initialization strategy for improving Fast AT? Following the same idea as in Section \ref{sec:method}, we tend to find an initialization $\bxi^*$ such that  
\begin{align*}
    \bxi^* = \argmin_{\bxi \in \cB_{\epsilon}(\zero)} ([f_{\bm\theta}(\xb + \bxi) - \gamma\bpsi]_y)^2,
\end{align*}
where $\bpsi$ is the random vector and we only take the $y$-logit since the CrossEntropy loss used in adversarial training mainly cares about the $y$-logit.
We test this on CIFAR-10 using ResNet-18 model, and summarize the results in Table \ref{table:pgd_backward}. We can observe that combining Backward Smoothing with PGD can still achieve certain level of improvements but not as good as when combined with TRADES.

\subsection{Comparison of Backward Smoothing and the ODI attack}
\citet{tashiro2020diversity} proposed an ODI attack which shares similar idea as our proposed Backward Smoothing method in terms of attack (it also computes an initialization direction before normal attack), however, we are not directly using ODI attack as our initialization. 
In this subsection, we compare the differences between them.
First, the formulation is different. In ODI attack, its initialization is solved by 
\begin{align*} 
    \max_{\bxi \in \cB_{\epsilon}(\bm{0})}  f_{\btheta} (\xb )^\top\bpsi,
\end{align*}
where $\bpsi \sim U(-1,1)$. This is a simple linear formulation compared to the KL divergence relationship \eqref{eq:output_smoothing_input} for Backward Smoothing. The use of KL divergence is actually quite natural in our case since TRADES method utilize KL loss for regularizing robust predictions.
Second, the motivations are different, ODI is a type of adversarial attack, which aims at lowering the prediction accuracy of target classifier, while our Backward Smoothing focuses on smoother the objective function. In fact, if one adopts Backward Smoothing for an attack, it can hardly achieve superior performances (smoother loss landscape also means hard to increase the loss significantly). To be more convincing, we also compare the result of applying ODI attack for adversarial training in Table \ref{table:pgd_backward}. We can observe that using ODI attack for robust training achieves much worse robustness performances compared to Backward Smoothing both for PGD based and TRADES based strategies.


\begin{table}[h!]

\caption{Performance of single-step based robust training strategy on CIFAR-10 dataset using ResNet-18 model.}
\label{table:pgd_backward}
\centering
\begin{small}
\begin{tabular}{lcc}
\multicolumn{1}{c}{\bf Method}  &\multicolumn{1}{c}{\bf Nat (\%)} &\multicolumn{1}{c}{\bf Rob (\%)}\\
\hline 
Fast AT & 84.79 & 46.30\\
Fast TRADES  & 84.80 & 46.25 \\
Backward Smoothing (PGD)  & 82.69 & 47.96\\
Backward Smoothing (TRADES) & 82.38 &
52.50\\
ODI (PGD) & 85.20 & 43.41 \\
ODI (TRADES) & 84.83 & 49.37
\end{tabular}
\end{small}
\end{table}


\end{document}